\newif\ifsubmit
\submitfalse   

\ifsubmit
  \documentclass[sigconf,anonymous,review]{acmart}
\else
  \documentclass[sigconf,nonacm]{acmart}
\fi

\setcopyright{none}                      
\ifsubmit
  \copyrightyear{}                        
\fi
\ifsubmit
  \acmConference[WWW 2026]{The ACM Web Conference 2026}{June 29--July 3, 2026}{Dubai, United Arab Emirates}%
\fi
\acmDOI{}
\acmISBN{}

\usepackage{booktabs}
\usepackage{multirow}
\usepackage{graphicx}
\usepackage{subcaption}
\usepackage{amsmath}
\usepackage[ruled,vlined]{algorithm2e}

\newcommand{\LatWeave}{\textsc{LatWeave}}
\newcommand{\meet}{\wedge}
\newcommand{\join}{\vee}
\newcommand{\leqpos}{\sqsubseteq}
\usepackage{xcolor}

\begin{document}

\title{Meet, Compare, or Abstain: \LatWeave{} for Deterministic Multi-Hop Question Answering on Knowledge Lattices}

\ifsubmit
\else
\author{Yuze Ren}
\affiliation{%
  \institution{ZenSmart Technology (Beijing) Co., Ltd., China}
  \country{}}
\email{renyuze@zhongshuruizhi.com}

\author{Shaoheng Fan}
\affiliation{%
  \institution{ZenSmart Technology (Beijing) Co., Ltd., China}
  \country{}}
\email{fanshaoheng@zhongshuruizhi.com}

\author{Tao Wang}
\affiliation{%
  \institution{ZenSmart Technology (Beijing) Co., Ltd., China}
  \country{}}
\email{wangtao@zhongshuruizhi.com}

\author{Yabo Yan}
\affiliation{%
  \institution{ZenSmart Technology (Beijing) Co., Ltd., China}
  \country{}}
\email{yanyabo@zhongshuruizhi.com}

\author{Han Han}
\authornote{Corresponding author.}
\affiliation{%
  \institution{ZenSmart Technology (Beijing) Co., Ltd., China}
  \country{}}
\email{hanhan1@zhongshuruizhi.com}
\fi

\begin{abstract}
Probabilistic question-answering systems---whether large language models (LLMs) themselves, retrieval-augmented generation (RAG), or trained multi-hop retrievers---conflate ``what is known'' and ``how to reason'' into a single probabilistic computation: hallucination cannot be eradicated, evidence chains cannot be audited, and the system answers even when it does not know. We present \LatWeave{}, which organizes knowledge into a multidimensional knowledge lattice and compiles multi-hop QA into three deterministic operators---meet (constraint intersection), compare (lattice-order comparison), and abstain (structural abstention); LLMs appear only on the construction side (one-shot extraction) and the query-planning side, while the answer-generation path is zero-LLM, zero-task-training, and auditable end to end---so that question answering over Web-published knowledge becomes reproducible item by item. Rather than claiming across-the-board SOTA, we characterize the operating envelope of this paradigm on six public benchmarks: when knowledge is complete (MetaQA, 39{,}093 questions) meet chains are near-lossless over three hops (any-hit 0.9975, on par with fully supervised KBQA); on templated multi-hop home ground (2WikiMultihopQA held-out $n=1{,}258$) EM 0.865, well above published structure-augmented RAG reproductions; on open-text deep composition (MuSiQue) and extraction-coverage gaps (HotpotQA) we report degradation honestly and attribute it to causes outside the lattice-algebra layer; and when information is incomplete (IIRC) we achieve structural abstention with abstain accuracy 0.971 and leak rate 0.029. Within the operating envelope, deterministic execution pays no performance penalty, and every step on the answer path can be recomputed---precisely the source of end-to-end auditability.
\end{abstract}



\maketitle


\section{Introduction}

Modern question answering systems have nearly converged on a hybrid architecture:
keyword retrieval guarantees exact matches, semantic retrieval handles recall, and
GraphRAG-style methods further introduce structured retrieval into the pipeline,
explicitly connecting the correlations that remain isolated between semantic chunks.
Engineering mechanisms such as chunk IDs have given the evidence chain a traceable
form in production for the first time. Yet two structural problems remain unsolved.
The first is hallucination: provenance annotations can trace which snippets an answer
consulted, but not whether the answer is logically entailed by them---the final step
from snippets to answer is still taken in probability space, and distortion,
extrapolation, and outright fabrication all occur at this step. This is not an
engineering oversight: hallucination is a mathematical inevitability of
probability-based generation---any calibrated language model must
hallucinate~\cite{kalai2024calibrated}. The second is answering anyway by design:
when neither retrieval nor the graph contains facts supporting an answer, the system
has no structural mechanism to stop itself.

The two problems trace back to the same root: engineering-level auditability---the
sources are traceable, but the reasoning from evidence to conclusion is a black box.
This paper argues for mechanism-level auditability: the answer itself
is a meet chain unfolded step by step, each step a verifiable lattice-algebra
operation that any examiner can replay and check independently. A clarification of
scope: this paper does not claim that deterministic methods outperform probabilistic
ones across all multi-hop tasks; what we characterize is where they approach
losslessness, where they structurally degrade, and how they abstain honestly when the
paradigm degrades---its \emph{operating envelope}. Mapping this envelope out is itself one of the
contributions of this paper.

\begin{figure*}[t]
  \centering
  \includegraphics[width=0.70\textwidth]{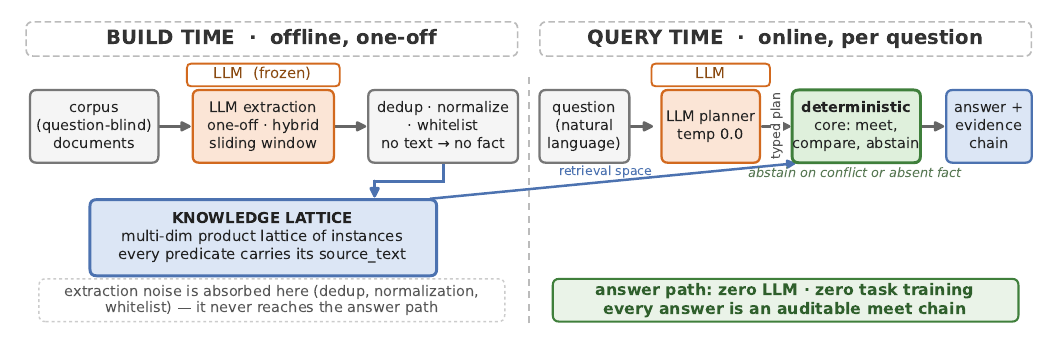}
  \caption{LatWeave in two stages. \textbf{Build time} (offline, one-off): an LLM
  extracts typed predicates from question-blind documents (dedup, normalization and a
  dimension whitelist absorb extraction uncertainty before it can enter the lattice),
  yielding a product lattice whose instances each carry their \texttt{source\_text}.
  \textbf{Query time} (online, per question): an LLM planner compiles the question
  into a typed plan, and a deterministic core executes the three operators
 ---\emph{meet}, \emph{compare}, \emph{abstain}. The LLM is confined to the two
  orange boxes (extraction and planning); the answer path carries zero LLM calls
  and zero task training, so every answer is an auditable meet chain.}
  \label{fig:overview}
\end{figure*}

Our approach can be summarized in one sentence: the LLM is restricted to a
``builder + query translator,'' and answers are produced deterministically by three
operators on the lattice. \emph{Meet} (constraint intersection) compiles the question
into a set of dimension constraints and takes their greatest lower bound (GLB) on
the product lattice; the hit instances form the answer, each carrying its
\texttt{source\_text}, so the evidence can be audited item by item. \emph{Compare}
(lattice-order comparison) performs comparison and set operations along the lattice
order---Boolean tests, complete answer-set return, and lattice-order adjudication of
``which is more $X$.'' \emph{Abstain} (structural refusal): when constraints
conflict or the fact is absent from the lattice, the system refuses
structurally---a direct consequence of lattice semantics, not a low-confidence
fallback. Figure~\ref{fig:overview} gives the two-stage overview: at build
time (offline, one-off) an LLM extracts typed predicates from question-blind
documents and constructs the lattice; at query time an LLM planner compiles
the question into a typed plan that the three operators execute
deterministically. LLM participation is confined to the builder and planner
roles---the answer path makes zero LLM calls and requires zero task
training---and every answer is an auditable meet chain.

Our contributions are fourfold:
\begin{itemize}
  \item \textbf{Mechanism.} A formalization and system realization of the knowledge
  lattice with the meet/compare/abstain operators; the answer path contains zero LLM
  calls and zero task training (\S3--4).
  \item \textbf{Upper-bound evidence.} With complete or templated knowledge,
  deterministic retrieval is near-lossless: MetaQA any-hit 0.9975 (zero training, on
  par with fully supervised KBQA) and 2Wiki held-out EM 0.865 (\S6).
  \item \textbf{Boundary evidence and attribution.} Degradation on three boundary
  benchmarks (MuSiQue / HotpotQA / FRAMES) is attributed layer by layer to causes
  outside lattice algebra, consolidating into a unified failure classification (\S7).
  \item \textbf{Abstention as a first-class capability.} A two-scale evaluation
  framework (recall $\times$ precision), corroborated on IIRC by structural
  abstention: abstain accuracy 0.971, leak rate 0.029, zero drift on the tuning
  pool (\S8).
\end{itemize}

The lineage of this work is also worth stating: \LatWeave{} builds on an
earlier rule-centric knowledge base system\ifsubmit\else~developed within our group\fi,
where partial ordering on a classification lattice served solely for
applicability gating, not yet a retrieval algebra; we generalize that gating into
a full multi-dimensional product-lattice algebra. The step from ``partial ordering
used only for gating'' to ``the lattice is the retrieval space'' is precisely where
the core contribution of this paper lies.

\section{Related Work}

Today's QA systems have evolved along four technical lines, each solving a different
problem: neural multi-hop retrievers achieve multi-hop recall but depend on model
training; RAG and GraphRAG mitigate knowledge staleness and chunk isolation, yet
hallucination cannot be eliminated and evidence chains remain hard to audit; KBQA
and semantic parsing deliver precise answers, but are bound to a predefined schema
and a trained parser; selective prediction lets a system refuse when it does not
know, but the refusal rests on a confidence threshold whose calibration drifts with
the task distribution. Only this work achieves zero training, deterministic
answers, mechanism-level auditability, and structural abstention at the same
time---these four properties have never before appeared in a single system
(Appendix Table~\ref{tab:positioning}).

\subsection{Multi-Hop Question Answering}
HotpotQA~\cite{yang2018hotpot}, 2WikiMultihopQA~\cite{ho2020constructing},
MuSiQue~\cite{trivedi2022musique}, IIRC~\cite{ferguson2020iirc}, and
FRAMES~\cite{krishna2025frames} together define the difficulty spectrum of
multi-hop QA---from templated two-hop questions to adversarial deep
composition, and from complete to incomplete information. Mainstream solvers
(MDR~\cite{xiong2021mdr}, IRCoT~\cite{trivedi2023ircot}, Beam
Retrieval~\cite{zhang2024beamretrieval}, HopRAG~\cite{liu2025hoprag}) are
probabilistic or training-dependent---a different precondition from our zero-training
setting.

\subsection{Retrieval-Augmented and Graph-Augmented LLMs}
RAG~\cite{lewis2020retrieval} and its refinements (Self-RAG~\cite{asai2024selfrag}
learns when retrieval pays off; FLARE~\cite{jiang2023flare} re-retrieves
proactively mid-generation) use semantic-similarity retrieval to mitigate
hallucination; GraphRAG~\cite{edge2024from}, LightRAG~\cite{guo2024lightrag},
HippoRAG~2~\cite{gutierrez2025hipporag2}, ToG~\cite{sun2024thinkongraph},
RoG~\cite{luo2024reasoning}, and KAG~\cite{liang2025kag} further introduce
graph structure into retrieval, mining the associations between semantic chunks.
This line strengthens ``retrieval,'' yet answers are still LLM-generated---the
hallucination and auditability problems remain exactly as before.

\subsection{KBQA and Semantic Parsing}
KBQA systems reason deterministically with fully verifiable
inference---KV-Mem~\cite{miller2016key}, GraftNet~\cite{sun2018open},
SRN~\cite{qiu2020stepwise}, PullNet~\cite{sun2019pullnet},
EmbedKGQA~\cite{saxena2020improving}, NSM~\cite{he2021improving}, and
RoG~\cite{luo2024reasoning}---but each depends on two prerequisites: a
ready-made knowledge graph and a trained parser. Our method carries neither:
a one-shot LLM extraction builds the lattice directly from text, keeping
auditability with no model-training requirement.

On the query side, the most recent publicly available work is LLM-Plan by
Shrestha and Kim~\cite{llmplanning2025kgqa}: an LLM plans only a relation
sequence that a deterministic BFS executes over a knowledge graph.
\LatWeave{} differs structurally: meet intersects per-dimension value
constraints where LLM-Plan's relation traversal carries none; we provide
compare semantics and structural abstention where LLM-Plan has neither; and we
characterize a six-benchmark operating envelope, while LLM-Plan is evaluated on
a single benchmark. The item-by-item numeric comparison appears in
Section~6.2.

\subsection{Abstention, Selective Prediction, and Hallucination}
Selective prediction supplies the theoretical frame for ``abstain when
unsure''~\cite{elyaniv2010foundations,geifman2017selective}, carried into
QA~\cite{kamath2020selective} and VQA~\cite{whitehead2022reliable};
TruthfulQA~\cite{lin2022truthfulqa} and hallucination
surveys~\cite{ji2023survey} chart the cost of never abstaining. Yet existing
abstention is almost entirely probabilistic confidence thresholding, whose
risk--coverage trade-off drifts with the task---softmax-score rejection
attains under 7.5\% coverage at 1\% risk, and a learned selector attains only about
15.6\%~\cite{whitehead2022reliable}. Our abstain is structural: a fact absent
from the lattice entails abstention, with no threshold to tune and a leak rate
that is empirically near zero (Section~\ref{sec:abstention}).

\subsection{Concurrent Work and Positioning}

Concurrent WWW 2026 work approaches different facets of the same problem:
S-Path-RAG~\cite{fu2026spathrag} trains a differentiable path scorer and injects
cross-attention into the LLM; CompactRAG~\cite{yang2026compactrag} cuts the
LLM-call and token cost of multi-hop QA to two calls per question, yet those calls
still synthesize the final answer; RoE~\cite{han2026reasoning} unifies
retrieval and generation via SFT- and RL-trained graph exploration;
HyperRAG~\cite{lien2026hyperrag} upgrades retrieval to n-ary hypergraphs with an
LLM-guided beam. We differ from all four on the same two poles: zero training,
and an answer path that never enters the LLM.
\section{The Knowledge Lattice Model}
\label{sec:model}

This section pins down the lattice as a mathematical object that a reviewer can
verify, and defines the three operators---\emph{meet} (constraint intersection),
\emph{compare} (lattice-order comparison) and \emph{abstain} (structural
refusal)---as algebraic operations with explicit semantics; the four classical
results the system puts to work, and the engineering question each answers, are
given at the end of Section~\ref{sec:posets}.

\subsection{Posets, Dimensions, and Product Lattices}
\label{sec:posets}

\begin{figure*}[t]
  \centering
  \includegraphics[width=0.75\textwidth]{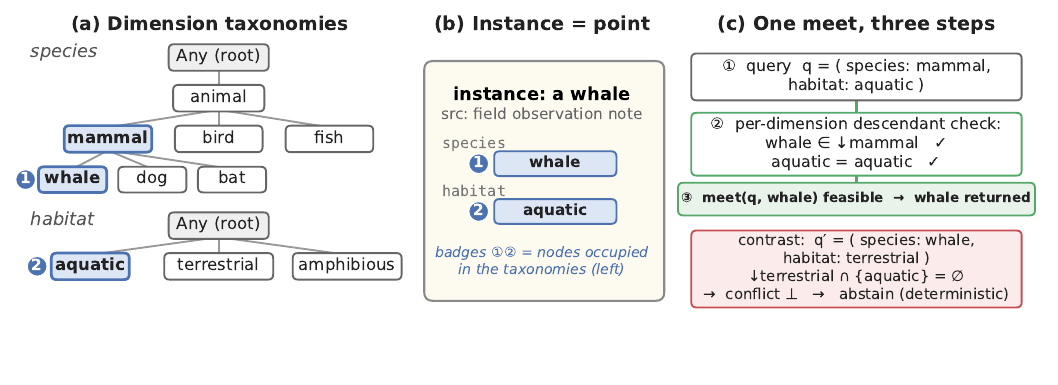}
  \caption{A product-lattice slice (species $\times$ habitat mini-example):
  (a) value hierarchies of the two dimensions; (b) an instance as a point in
  the product lattice; (c) one \emph{meet} in three deterministic steps---a hit
  requires the descendant check on every dimension, and a conflict on any
  dimension entails abstain.}
  \label{fig:lattice}
\end{figure*}

Start with a miniature example (Figure~\ref{fig:lattice}). Take two dimensions:
\emph{species}, a biological taxonomy (under \emph{animal} sit
\emph{mammal}, \emph{bird}, and \emph{fish}, and under \emph{mammal} sit
\emph{whale}, \emph{dog}, and \emph{bat}), and \emph{habitat}, whose values
include \emph{aquatic}, \emph{terrestrial} and \emph{amphibious}. An
instance---one particular whale---is bound to the product-lattice coordinate
$(\textit{whale}, \textit{aquatic})$. The query ``aquatic mammals'' then amounts
to a single \emph{meet}: \emph{whale} is a descendant of \emph{mammal} on the
first dimension and \emph{aquatic} matches the second, so the whale is returned.
If instead we ask for ``terrestrial whales,'' the two constraints conflict, so
the system abstains deterministically rather than guessing. The structure is no
toy: the same product-lattice slice already runs over \ifsubmit\else our group's\fi
oil and gas production corpus.

Three standard notions---\emph{poset} (a reflexive, antisymmetric, transitive
relation $\leqpos$ on a set $P$), \emph{lattice} (a poset in which every pair
$a,b$ has a greatest lower bound $a \meet b$ and a least upper bound
$a \join b$), and \emph{product lattice} (componentwise order, meet and join
over $L_1 \times \cdots \times L_k$)---freeze the intuition's left half
(see, e.g.,~\cite{davey2002introduction}); the two definitions and one
convention below are our own, attaching dimensions and instances to the
product lattice and covering instances that miss a dimension.

\textbf{Definition 4 (Dimension).} Each knowledge dimension $L_i$ is a rooted
directed acyclic graph: its nodes are dimension values (including synonym sets
and numeric intervals), its edges encode the ``more general $\to$ more
specific'' order, and its root is $\top$.

\textbf{Definition 5 (Instance).} An instance is bound to a coordinate of the
product lattice; its predicate set $\{(d_i, v_i, \tau_i)\}$ gives, for each
dimension, a value and a confidence. Every predicate must carry its
\texttt{source\_text}---an anti-hallucination rule: an assertion without
supporting evidence text is not admitted into the lattice.

\textbf{Convention (embedding of missing components).} When an instance misses
a dimension, that component is read as $\top$ (the most general value) and
violates no order check. A ``query'' is then stated precisely as a meet on the
product lattice---the subject of the next subsection.

Four classical results are put to work by the system, each answering one
engineering question.
(1)~\emph{Is the lattice legitimate?} Birkhoff's distributivity identity
$x \meet (y \join z) = (x \meet y) \join (x \meet z)$ turns the naive
$M_3/N_5$ witness search into a per-triple check ($O(n^3)$ with bitset
meet/join tables)~\cite{davey2002introduction}; dimensions containing
ambiguous sublattices never go live.
(2)~\emph{Can we afford meet?} On-demand Dedekind--MacNeille completion
computes GLBs by bitset intersection plus extremum extraction, replacing the
$2^{|P|}$ enumeration.
(3)~\emph{How is cross-dimension structure discovered?} Formal Concept
Analysis (FCA) identifies attributes shared across dimensions and supports
cross-reference
expansion~\cite{wille1982restructuring,ganter2012formal}.
(4)~\emph{Are the dimensions independent?} The Dilworth width, obtained via
K\H{o}nig bipartite matching, assesses the mutual independence of
dimensions~\cite{dilworth1950}.

\subsection{Three Deterministic Operators and the Capability Boundary}
\label{sec:operators}

\paragraph{meet (constraint intersection)} The semantics of a constraint set
$C = \{(d_i, V_i)\}$ is per-dimension GLB followed by coordinate composition
in the product lattice: for every constrained dimension we intersect the
descendant sets of the candidate values and extract the greatest element; a
dimension with no common lower bound is a conflict in that dimension. A
conflict is not an exception but a first-class output of the lattice
semantics---it is the legitimate trigger of \emph{abstain}.

\paragraph{compare (lattice-order comparison)} One operator, four
instantiations on the lattice order: (1)~Boolean comparison of two entities
along the same dimension (2Wiki comparison questions); (2)~order-based
selection of an extremum among candidates (bridge\_comparison's ``which is
more $X$''); (3)~complete-set return---\emph{meet} returns the full answer
set rather than a top-1 ranking (the primary MetaQA metric,
Table~\ref{tab:metaqa_main}); (4)~Boolean boundary---when the compared
objects are not in the lattice, \emph{compare} degrades to \emph{abstain}
(IIRC binary questions, Table~\ref{tab:iirc_attribution}). The four forms
share one algebraic basis: instantiations of a single operator, not four
patched-together features.

\paragraph{abstain (structural abstention)} A monotone abstention semantics:
if the constraints conflict, or no lattice point satisfies all constraints
(the hit set is empty), the system
abstains. The tighter the constraints, the weaker the answerability---never
the reverse.

\paragraph{Capability boundary (declared a priori)} Any question that derives
new values on top of lattice values (arithmetic, derived counts) belongs to
external operator modules, which this paper neither implements nor counts in
the in-scope numbers. This declaration licenses the FRAMES capability strata
(Appendix~\ref{app:frames}) and the IIRC value-type accounting
(Table~\ref{tab:iirc_attribution}).

Three short propositions follow, each with a one-sentence proof idea (the full
arguments are immediately checkable).
\emph{Proposition 1 (determinism):} the same lattice and the same query yield
the same result, because GLB and lattice-order comparison are functions with
no random source.
\emph{Proposition 2 (auditability):} every answer unfolds into a \emph{meet}
chain, and every instance on the chain carries its \texttt{source\_text};
auditing means checking the algebra and the evidence text step by step.
\emph{Proposition 3 (monotone abstention):} tightening the constraints
(adding a dimension or taking a more specific value) never turns an
\emph{abstain} into an answer, because tightening shrinks the satisfying set
monotonically and an empty set stays empty.


\subsection{Relation to FCA and Knowledge Graphs}
We borrow lattice semantics from FCA, but our objects are multi-attribute
instances rather than formal concepts; we borrow entity-relation structure
from knowledge graphs (KGs), but keep dimensional coordinates and the lattice
order, making \emph{meet}/\emph{compare} native operations rather than
bolted-on traversals. FCA answers ``which objects share which attributes'',
whereas we answer ``given constraints, which lattice points satisfy them; and
when none do, is it a conflict or a gap''.
\section{\LatWeave{}: System Design}

This section presents the system design of \LatWeave{}. At construction time
(Section~\ref{sec:construction}) open text is woven into a knowledge lattice;
at query time (Section~\ref{sec:query}) deterministic three-state execution
runs on the lattice; Section~\ref{sec:impl} closes with implementation and
efficiency notes. The LLM appears in exactly two places---extraction at
construction time and planning at query time---and its uncertainty is kept
off the answer path. Engineering details are limited to feasibility numbers;
the rest goes to the appendix.

\subsection{Construction: From Text to Lattice}
\label{sec:construction}

Construction turns open text into a knowledge lattice; the core is hybrid
extraction. Short documents go through the whole-doc mode, where the full
text is fed to the LLM once; long documents go through a physical sliding
window cut by character position (window size derived from the model's
context size, consecutive chunks overlapping 200--500 characters), so that
no content is lost during chunking. Extracted facts then pass a strict
grid-point hash dedup: a fact repeated by the LLM---same lattice coordinates,
same source evidence---enters the lattice only once. The last step is
whitelist-guided extraction with dimension normalization: the prompt carries
a table of preferred dimensions whose canonical names are used when they
match semantically, uncovered relations may be named freely, and a variant
table normalizes synonymous variants afterward; LLM-invented dimensions are
thereby mostly long-tail and low-frequency. The motivation is empirical: in
MuSiQue construction, free naming once shattered the single relation
``educated at'' into 1{,}017 dimension names.

\subsection{Query Time: Planning and Deterministic Three-State Execution}
\label{sec:query}

Query time has two parts: the LLM planner compiles the question into a typed
plan, and deterministic execution runs that plan on the lattice and returns
one of three states. The planner is a replaceable semantic-parsing front end
(temperature $0.0$; plans are cacheable and reproducible), and its loss can
be measured separately from lattice competence by an oracle-plan ceiling
experiment (Section~\ref{sec:analysis}). For ``physicists born in Vienna who
won the Nobel Prize in Physics'' the compiled plan is a conjunction of three
dimension meets, and the executor intersects them to obtain the complete
answer set (a \emph{meet} hit); if the constraints conflict or the hit set is
empty, it abstains---the refusal comes directly from on-lattice execution,
with no external fallback of any kind; \emph{compare} returns a Boolean, an
extremum, or a complete-set verdict. Union semantics applies to the
\emph{meet} path only, because complete-set semantics must not be diluted by
unions. All experiments in this paper evaluate exactly this three-state
execution. The executor itself is three lines of logic: per-dimension
descendant-set intersection, extremum extraction, and conflict return when a
dimension has no common lower bound.


\subsection{Implementation and Efficiency}
\label{sec:impl}

This section gives feasibility evidence: the system runs at realistic scale,
and it is fast. Scale: the 2Wiki dataset holds 1.2M instances, built from
the full dev corpus of 125{,}760 passages. Efficiency: lexicon-scan
construction takes 235\,ms; retrieval executes at p50 below 18.7\,ms and
p95 below 50\,ms; planning for the 1{,}258 held-out questions replays from
cache in about 1\,s. Against GraphRAG-style systems: their indexing requires
multiple LLM passes over the whole corpus plus maintained embeddings, and an
LLM still sits in the query loop---the gap is architectural, not something
that engineering-level tuning can close.
\section{Experimental Setup}
\label{sec:setup}

\subsection{Datasets and Their Roles}

We evaluate \LatWeave{}'s retrieval on six public datasets, each serving one
role in our argument (Table~\ref{tab:datasets}). MetaQA is the control:
with knowledge complete, it verifies that the lattice algebra itself loses
nothing. 2WikiMultihopQA is the main experiment. HotpotQA and MuSiQue are
the two boundary experiments, exposing respectively the fact-coverage gap of
open-domain extraction and the degradation as chains lengthen (2$\to$4 hops).
IIRC examines refusal under incomplete information. FRAMES is a compounded
stress test (Appendix~\ref{app:frames}): fact-coverage gap $\times$ query
grounding.

\begin{table}[t]
  \centering
  \caption{Datasets, structure, and role in this paper's argument.}
  \label{tab:datasets}
  \footnotesize
  \setlength{\tabcolsep}{3pt}
  \begin{tabular}{@{}p{2.0cm}p{1.5cm}p{2.2cm}p{2.0cm}@{}}
    \toprule
    Dataset & Size used & Structure & Role \\
    \midrule
    MetaQA \cite{zhang2018variational} & test $n=39{,}093$ & 1/2/3-hop: 9{,}947 / 14{,}872 / 14{,}274 & Control: complete knowledge $\Rightarrow$ lattice algebra lossless \\
    2WikiMultihopQA \cite{ho2020constructing} & held-out $n=1{,}258$ & 4 question types & Main experiment: templated multi-hop, end-to-end \\
    HotpotQA \cite{yang2018hotpot} & held-out $n=741$ & bridge 592 / comparison 149 & Boundary 1: fact-coverage gap \\
    MuSiQue \cite{trivedi2022musique} & held-out $n=242$ & 2/3/4-hop: 125/76/41 & Boundary 2: longer-chain degradation \\
    IIRC \cite{ferguson2020iirc} & held-out $n=130$ & span 59/ value 23/ binary 13/ none 35 & Refusal experiment: abstain under incompleteness \\
    FRAMES \cite{krishna2025frames} & dev 742 + held-out 82 & factuality $\times$ multi-hop & Compounded stress test: coverage $\times$ grounding \\
    \bottomrule
  \end{tabular}
\end{table}

\subsection{Split Rules and Leakage Control}

All text benchmarks share one split rule: a stratified 90/10 dev/held-out
division (seed 88). Lattice construction is strictly question-blind---only
paragraphs or KB triples are read, never questions or answers.
Configurations are frozen on dev, and the held-out split is evaluated
exactly once. MetaQA keeps its official test split (39{,}093 questions); its
configuration is frozen on the official dev split with a 9-shot planner
prompt (exemplars drawn only from the train split, so no test information
leaks). Held-out drift is zero: 2Wiki 0.886 vs.\ 0.865; IIRC precision
measure 0.9682 vs.\ 0.9437.

\subsection{Baselines and Condition Labeling}

Baselines fall into two blocks. The \emph{zero-training block} runs on
the same corpus and the same scoring function as ours: a keyword baseline,
Naive RAG (TF-IDF retrieval plus an LLM reader), and LatWeave. The
\emph{reference block} reports published values with protocol differences
footnoted (GraphRAG 0.514$^{\ddagger}$, HippoRAG~2 0.650$^{\ddagger}$, IRCoT
0.53, HopRAG 0.62, and the supervised family on MetaQA); it serves as a
paradigm reference only, and we draw no same-condition win/loss verdicts
from it.

\subsection{Metrics}
\label{sec:metrics}

Our metric family is organized around three abilities: \emph{answering
correctly} (EM / F1), \emph{answering completely} (any-hit / complete-set
recall), and \emph{refusing honestly} (Faithfulness / dual measures / leak
rate). EM and F1 come from a single scoring function: prediction and gold
pass through the same text normalization; EM tests exact equality of the
normalized strings, and F1 is token-level overlap. Since EM${=}1$ forces
F1${=}1.0$, our F1 convention is equivalent to $\mathrm{F1}^{*} :=
\max(\mathrm{F1}_{\mathrm{raw}}, \mathrm{EM})$. Every LLM call is pinned to
temperature 0.0 and seed 88; the FRAMES capability classes are assigned by a
frozen mechanical decision table whose rules read only the question and the
gold answer---never the model output (Appendix~\ref{app:protocol}).
\section{Main Results: Meet at Its Best}

\subsection{2WikiMultihopQA: The Main Battlefield}

\begin{table}[t]
  \centering
  \caption{2WikiMultihopQA main results (held-out $n=1{,}258$).}
  \label{tab:twowiki_main}
  \small
  \begin{tabular}{p{3.4cm}ccc}
    \toprule
    Method & EM & F1* & Resolved \\
    \midrule
    Keyword baseline & 0.083 & 0.111 & 73.2\% \\
    Naive RAG (TF-IDF + LLM) & 0.198 & 0.205 & 100\% \\
    GraphRAG \cite{edge2024from} & 0.514$^\ddagger$ & 0.586$^\ddagger$ &---\\
    HippoRAG 2 \cite{gutierrez2025hipporag2} & 0.650$^\ddagger$ & 0.710$^\ddagger$ &---\\
    IRCoT \cite{trivedi2023ircot} & 0.53 & 0.65 &---\\
    HopRAG \cite{liu2025hoprag} & 0.62 & 0.69 &---\\
    Beam Retrieval \cite{zhang2024beamretrieval} &---&---&---\\
    OPEN-RAG \cite{islam2024openrag} &---&---&---\\
    Self-RAG \cite{asai2024selfrag} &---&---&---\\
    \midrule
    \LatWeave{} (held-out) & \textbf{0.865} & \textbf{0.872} & 92.2\% \\
    \quad (tune pool $n=10{,}179$, reference) & 0.886 &---& 92.9\% \\
    \bottomrule
  \end{tabular}
  \vspace{1pt}
  \par\footnotesize\raggedright
  $^\ddagger$~Published values~\cite{ko2025cooprag} carry protocol differences
  (1k-question subset, restricted corpus, Llama-3.3-70B reader): a paradigm
  reference, not a same-condition comparison. Beam Retrieval, OPEN-RAG, and
  Self-RAG are trained methods, listed for positioning without numbers.
\end{table}

\begin{table}[t]
  \centering
  \caption{2WikiMultihopQA per-type breakdown (held-out $n=1{,}258$).}
  \label{tab:twowiki_types}
  \small
  \begin{tabular}{llll}
    \toprule
    Question type & Share & EM & $\Delta$ vs.\ overall \\
    \midrule
    bridge\_comparison & 21.9\% & 0.975 & +11.0pp \\
    comparison         & 24.2\% & 0.888 & +2.3pp \\
    compositional      & 41.7\% & 0.815 & $-$5.0pp \\
    inference          & 12.3\% & 0.794 & $-$7.1pp \\
    \midrule
    Overall            & 100\%  & 0.865 &---\\
    \bottomrule
  \end{tabular}
\end{table}

The breakdown brings the compare family to numbers for the first time:
bridge\_comparison (0.975) and comparison (0.888) lead, and these two are
precisely the compare operator's home ground (Section~\ref{sec:operators})
--- lattice-order side selection and Boolean comparison, where lattice
comparability yields the answer directly with no generation.

On held-out ($n=1{,}258$) \LatWeave{} reaches EM 0.865, F1* 0.872, and
resolved 92.2\%, and on the resolved questions EM climbs to 0.938: on the
questions it answers, the system is 93.8\% precise. The relative gains over the same-protocol Naive RAG rerun (0.198) and the
strongest published graph-RAG value (HippoRAG~2, 0.650) are 337\% and 33\%
respectively. The tune pool
($n=10{,}179$, disjoint from held-out) gives EM 0.886, 2.1pp above held-out
--- tuning did not overfit the held-out set. Ablation (Appendix
Table~\ref{tab:twowiki_ablation}) shows that from the dev-only baseline 0.731
to the frozen configuration the largest single increment comes from train-set
transfer (+9.5pp), with alias normalization (+2.0pp) and rule-based
re-extraction (+1.0pp) accounting for the rest.

\subsection{MetaQA: The Representation-Independence Controlled Comparison}
MetaQA is a templated QA dataset: every question is template-generated and
every answer is drawn strictly from the knowledge graph, so on the same
questions any hit-rate gap between the knowledge-graph and knowledge-lattice
representations is fully attributable to the representation form itself. The
most recent publicly available work on the same benchmark is LLM-Plan by Shrestha and
Kim~\cite{llmplanning2025kgqa}, evaluated on the identical
39{,}093-question test split (Table~\ref{tab:metaqa_sota}).
The 2Wiki results above rest on LLM open extraction, so a natural objection is
whether the performance comes from the extraction or from the lattice algebra
itself. MetaQA is a purely structured KBQA benchmark (134{,}741 triples /
43{,}234 entities / 9 relations) whose triples map to lattice points with zero
LLM---dimensions are the relations (9 forward + 9 inverse + 1 subject anchor),
instances the subject- and object-side entities ($269{,}482$)---so the same
engine runs losslessly when the knowledge source is swapped from text to KB,
turning ``the lattice is a general knowledge representation, not a byproduct
of LLM extraction'' into a falsifiable claim. KB coverage and seed-entity
linkage are both 100\% (measured), so the only source of loss is question
parsing.

The evaluation setup mirrors the 2Wiki experiment: question-blind
construction (minutes on \texttt{kb.txt}); configuration frozen on dev
(9-shot, temp 0.0 / seed 88); test evaluated once. Parsing costs one LLM
call per question (parse\_fail = 0); query execution is a pure
meet-algebra chain with zero LLM.

\begin{table}[t]
  \centering
  \caption{MetaQA test main results ($n=39{,}093$, single final run, frozen 9-shot configuration).}
  \label{tab:metaqa_main}
  \small
  \begin{tabular}{llllll}
    \toprule
    Hop & $n$ & any-hit & recall (complete) & EM (strict) & set-F1 \\
    \midrule
    1-hop & 9{,}947  & 0.9990 & 0.9987 & 0.9447 & 0.9798 \\
    2-hop & 14{,}872 & 0.9983 & 0.9983 & 0.6239 & 0.9448 \\
    3-hop & 14{,}274 & 0.9955 & 0.9954 & 0.5304 & 0.9369 \\
    \midrule
    All   & 39{,}093 & \textbf{0.9975} & \textbf{0.9973} & 0.6714 & 0.9508 \\
    \bottomrule
  \end{tabular}
\end{table}

\begin{table}[t]
  \centering
  \caption{MetaQA test vs.\ published KBQA methods (supervised rows: Hits@1
  verified from source papers, converted to decimals). The Shrestha--Kim row
  reports Hit Rate---verbatim the any-hit measure, hence directly comparable
  to our any-hit row; their pooled micro-F1 is not cross-compared with our
  per-question set-F1 0.9508.}
  \label{tab:metaqa_sota}
  \footnotesize
  \setlength{\tabcolsep}{3pt}
  \begin{tabular}{@{}p{1.9cm}cccp{2.0cm}@{}}
    \toprule
    Method & 1-hop & 2-hop & 3-hop & Paradigm (trained) \\
    \midrule
    KV-Mem \cite{miller2016key}      & 0.962 & 0.827 & 0.489 & Memory networks (yes) \\
    GraftNet \cite{sun2018open}      & 0.970 & 0.948 & 0.777 & Subgraph retrieval (yes) \\
    SRN \cite{qiu2020stepwise}       & 0.970 & 0.951 & 0.752 & Neural reasoning (yes) \\
    PullNet \cite{sun2019pullnet}    & 0.970 & 0.999 & 0.914 & Graph retrieval (yes) \\
    EmbedKGQA \cite{saxena2020improving} & 0.975 & 0.988 & 0.948 & KGE (yes) \\
    NSM \cite{he2021improving}       & 0.971 & 0.999 & 0.989 & State machine (yes) \\
    RoG \cite{luo2024reasoning}      &---  &---  & 0.848 & LLM+graph (1k fine-tune) \\
    LLM-Plan + BFS \cite{llmplanning2025kgqa} & 0.999 & 0.999 & 0.970 & LLM plan + BFS (no task training) \\
    \midrule
    \LatWeave{} (any-hit)            & \textbf{0.9990} & \textbf{0.9983} & \textbf{0.9955} & Lattice meet (zero training) \\
    \LatWeave{} (recall, complete)   & 0.9987 & 0.9983 & 0.9954 & gold $\subseteq$ returned set \\
    \bottomrule
  \end{tabular}
\end{table}


Three observations. (i)~With no training, \LatWeave{} reaches hits comparable
to those of fully supervised methods: 1-hop (0.9990) and 3-hop (0.9955)
match or exceed the best published values, while 2-hop
(0.9983) comes within 0.07pp of the best (0.999). (ii)~The hop-decay profiles are categorically
different: KV-Mem degrades sharply with depth ($0.962\to0.489$), even the
strongest NSM shows visible 3-hop decay ($0.999\to0.989$), whereas
\LatWeave{} is almost flat across three hops (total span 0.0035)---hop
depth is not a difficulty variable for the lattice algebra. (iii)~Every
supervised baseline needs large-scale training (96k--330k questions), while
\LatWeave{} only reads the KB. Dev/test are highly symmetric (per-hop any-hit
deviation $\leq 8\times10^{-4}$); empty $= 69$ (0.18\%).

\textbf{Completeness is the compare family's second home ground:} meet returns
the full answer set deterministically, so completeness is a direct product of
the semantics---ranking-style methods, which treat ``gold enters the top-$k$
candidates'' as the hit endpoint, cannot structurally promise it. Our any-hit measure
--- 1 iff $\hat{A} \cap A^* \neq \emptyset$---is the Hits@1 of
the same returned set under an optimal internal ordering, hence an upper
bound on Hits@1; with set-F1 0.9508 the actual gap is negligible. Tightness
matters equally: on the single-answer subset ($|A^*| = 1$, $n = 14{,}834$)
any-hit and ranked Hits@1 are strictly equivalent, measured 0.9974.

Appendix Table~\ref{tab:metaqa_buckets} shows that recall (complete)---1 iff $A^*
\subseteq \hat{A}$---is $\geq 0.996$ in every answer-set-size bucket. Strict
answer-set EM ($\hat{A} \equiv A^*$) is 0.6714 and decays with answer-set size
(0.833 at $|A^*|=1$, 0.379 at $|A^*|\geq 11$)---an artifact of set strictness:
over 99\% of EM misses still intersect the gold set, no under-answering miss
exists, and all misses come from one or two extra or wrong entities.

The MetaQA control arm yields a direct production-route corollary: domains
that already hold structured knowledge (KG / database / table) should map it
straight to a lattice and skip LLM extraction entirely---zero training, zero
fine-tuning, any-hit 0.9975, sub-50\,ms retrieval, zero LLM. Three premises
are stated honestly: question parsing remains the only source of loss; fan-out
value selection still needs deterministic post-processing or a human in the
loop; and the lattice inherits the KG's missing edges and dirty triples
losslessly---input quality is the ceiling.
\section{Analysis: The Operating Envelope}
\label{sec:analysis}

This section turns the ``operating envelope'' from a position statement into a
quantified characterization. The claim: on the three multi-hop boundary
datasets (HotpotQA, MuSiQue; FRAMES in Appendix~\ref{app:frames}), all losses
fall outside the lattice-algebra layer, in two upstream layers---(a)
open-text-to-structure conversion (coverage gaps and dimension fragmentation
after whitelist governance; Sections~\ref{sec:construction}
and~\ref{sec:grounding}); (b) natural-language-to-lattice-query parsing
(query grounding and chain landing-point precision).

\subsection{Failure Attribution: Four Boundary Classes}
\label{sec:attribution}

On the three multi-hop boundary datasets, the dominant cause of lattice-retrieval
failure differs by class. On HotpotQA, the answer never enters the candidate
set: 90.6\% of residual broken chains lack the final-answer edge in the lattice
(Appendix Table~\ref{tab:attribution}), and among resolved-but-wrong questions,
77\% have their gold answer outside the returned set altogether; nor is the gap
unrepaired---build-layer repair lifts the formerly hop-1-broken bucket from EM
0 to 17.3\%, but conversion saturates and oracle injection falsifies
same-mechanism repair (Section~\ref{sec:grounding}). MuSiQue's attribution is
compositional decay across hops: 48.3\% of chains break at the first hop and a
single failing hop fails the whole chain (EM 0.0248). FRAMES compounds the
two. IIRC adds a fourth class---the answerability boundary
(Section~\ref{sec:boundaries}): answers must be derived on top of lattice
values (time differences, counts), which exceeds the declared deterministic
capability (Section~\ref{sec:operators}); or the compared subject is absent
from the lattice altogether---in both cases the correct behavior is to abstain
rather than guess.

\subsection{Hop Decay in Two Contrasts: Complete KB vs.\ Open Text}
\label{sec:hopdecay}

\begin{figure}[t]
  \centering
  \includegraphics[width=0.62\linewidth]{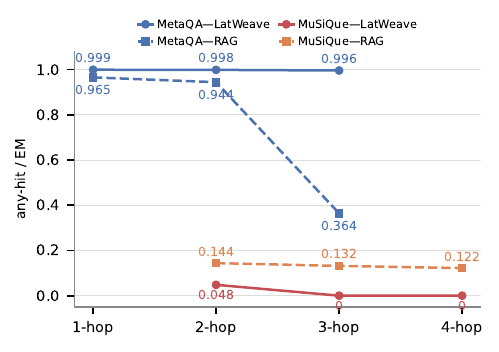}
  \caption{Hop decay in two contrasts. MetaQA (complete KB, any-hit): lattice meet
  chains are nearly lossless over three hops (0.9990 / 0.9983 / 0.9955), while the
  same-corpus RAG baseline (hits@1) decays to 0.364 at 3-hop. MuSiQue (open-text
  extraction, EM): lattice retrieval collapses to zero on multi-hop text composition
  (0.048 / 0 / 0), while RAG stays flat at a low 0.12--0.14 (insensitive to hop
  count).}
  \label{fig:hopdecay}
\end{figure}

The dual-contrast figure (Figure~\ref{fig:hopdecay}) shows jointly that hop depth
is not a difficulty variable for the lattice algebra itself; the degradation
comes from the knowledge source, not the reasoning core. Corroborating
per-hop conditional meet accuracy (Appendix Figure~\ref{fig:hopwise}, MuSiQue
tune pool $n=2{,}175$): 0.37--0.44 at hop 1, 0.09--0.22 at hop 2, approaching
zero at hop 3---each additional hop compounds conditional success
approximately multiplicatively, and the first hop is the largest loss surface.

The RAG contrast: on MuSiQue held-out, RAG reaches EM 0.1364 above the
lattice's 0.0248; on HotpotQA held-out, EM 0.4683 above 0.1687---the
coverage gap puts lattice retrieval's
full-pool EM at about one third of the same-corpus generative RAG baseline.
Once facts are inside, lattice retrieval itself is accurate (resolved-EM
0.4545 on HotpotQA, 0.938 on 2Wiki). MuSiQue falls into three classes by break
type: operational (OP\_final, $n=61$, EM 0.082); honestly resolvable
(QU\_honest, $n=28$, EM 0.0357); and structural-abstention (QU\_not\_honest,
anchor unresolvable, empty returned by design, $n=153$, EM 0)---all 6 correct
answers are 2-hop verbatim matches. On
HotpotQA, comparison questions resolve at 70.1\% while multi-hop meet
questions resolve at only 29.6\%: multi-hop meet must ground hop by hop,
compounding per-hop in-lattice rates and landing-point errors
multiplicatively.

\subsection{Query-Grounding Levels and Loss Upper Bounds}
\label{sec:grounding}

The oracle-plan upper bound separates planner loss from lattice capability in
numbers: on 2Wiki, resolved-EM is 0.938 (vs.\ 0.865 full-pool) and on HotpotQA
0.4545 (vs.\ 0.1687)---given successful parsing and chaining, on-lattice
execution is far above the end-to-end numbers. Appendix
Table~\ref{tab:attribution} puts the planner-attributable share of HotpotQA's
residual broken chains at 9.4\%, and FRAMES' oracle ceiling is only 4.9\%
(vs.\ EM 2.56\%)---on neither boundary is the planner the main bottleneck, so
repair investment shifts to the build layer (measured limits below).

Query grounding refers to where the planner's relation names come from when it lays
out a chain---generated from thin air, chosen from a global vocabulary, or chosen
from an entity's in-lattice relation menu (Table~\ref{tab:grounding}). We have
internalized L1 and L2 in the two boundary experiments respectively: the HotpotQA
planner is injected with the global relation vocabulary; the FRAMES planner first
queries which relations the seed entity actually has in the lattice and is
restricted to that menu---relation names are thereby anchored to edges that
genuinely exist on that entity. Yet both experiments show that L1/L2 remain
insufficient, for structural reasons.

\begin{table}[t]
  \centering
  \caption{Four levels of query grounding.}
  \label{tab:grounding}
  \small
  \setlength{\tabcolsep}{3pt}
  \begin{tabular}{@{}llp{2.0cm}p{2.4cm}@{}}
    \toprule
    Level & Mechanism & Grounding object & Status \\
    \midrule
    L0 & Free generation & none & Rejected \\
    L1 & Global vocabulary & global relation names & Internalized (HotpotQA) \\
    L2 & Entity menu & seed entity's lattice neighborhood & Internalized (FRAMES) \\
    L3 & Per-hop replanning & intermediate entities & Not adopted (argued below) \\
    \bottomrule
  \end{tabular}
\end{table}

The structural limitation is a circular dependency: static planning can, by
construction, ground only the first hop---the intermediate entity of hop-$k$
($k \geq 2$) is unknown until hop-$(k{-}1)$ has executed, so its dimension
menu is unavailable at planning time. The FRAMES chain-break dissection shows
this directly: an intermediate entity typically already carries 64--129
dimensions in the lattice, yet the break-point relation (e.g.,
\texttt{nominate\_for}) is still absent---the break is a chain-design error
(planning a relation the entity does not have), against which re-extraction
is ineffective in principle: exactly the position L2 cannot reach and only
L3 could.

Why not adopt L3? Three lines of experimental evidence. (1) FRAMES has internalized
L2, yet EM is still only 2.56\%; a per-hop replanning pilot hits 0 of 15
questions---the intermediate entities of broken chains are simply not in the
lattice (12 of 13 dissected cases), so replanning has nothing to choose from.
(2) HotpotQA hop-2 oracle injection gives L3's capability upper bound: placing
the gold answer directly into the correct hop-2 dimension still fails to
converge on 11 of 13 questions---the residual is an upstream coverage /
landing-point problem that no L3-level improvement can address. (3) L3 requires one LLM call
per hop, degrading the system to IRCoT-style iterative retrieval, which
is incompatible with this paper's ``zero LLM on the answer path'' positioning.

HotpotQA re-extraction, measured (details in Appendix~\ref{app:reextract}):
replaying all formerly hop-1-broken questions ($n=1{,}509$) under the frozen
repaired configuration lifts first-hop connectivity to 78.9\%, but only
21.9\% of connected questions are ultimately answered correctly (bucket EM
0 $\to$ 17.3\%); extrapolated to all first-hop gaps, full-pool EM gains only
about +0.6pp. A per-question residual decomposition rewrites the bottleneck
from ``choosing wrong among candidates'' to ``gold not in the returned set''
(77\% of resolved-but-wrong questions; returned sets average just 3.6
candidates). Fundamental improvement requires changing the extraction
mechanism itself; the IIRC value-cleaning repair
(Section~\ref{sec:ablation}) has not been applied here (future work).
Table~\ref{tab:envelope} summarizes the operating envelope with frozen
final values.

\begin{table}[t]
  \centering
  \caption{Operating envelope (all numbers are frozen final values).}
  \label{tab:envelope}
  \footnotesize
  \setlength{\tabcolsep}{3pt}
  \begin{tabular}{@{}p{1.7cm}p{2.2cm}p{2.6cm}p{1.3cm}@{}}
    \toprule
    Dataset & Knowledge-source form & Main bottleneck & Result \\
    \midrule
    2WikiMultihopQA & Open text extraction, templated questions &---(home ground) & EM 0.865 \\
    MetaQA & KB triples mapped directly & Question parsing only & any-hit 0.9975 \\
    HotpotQA & Open text extraction, free paragraphs & Coverage gap (90.6\%) & EM 0.1687 \\
    MuSiQue & Open text extraction, 4-hop chains & Chain deepening (48.3\% hop-1 break) & EM 0.0248 \\
    FRAMES (dev $n=742$) & Open text extraction, multi-hop $+$ factuality & Coverage gap $+$ chain-design error & EM 0.0256 \\
    IIRC (held-out $n=130$) & Open text extraction, incomplete information & Answerability boundary & Prec.\ measure 0.9437 \\
    \bottomrule
  \end{tabular}
\end{table}
\section{Abstention as a First-Class Capability}
\label{sec:abstention}

\subsection{IIRC: Abstention in Practice}

Traditional EM/F1 mixes questions on which the system should abstain with the
answerable ones, hiding the cost of bluffing. We separate a \emph{recall
measure} (EM on answerable questions) from a \emph{precision measure} (abstain
accuracy $\times$ (1 $-$ leak rate)), making ``knowing when not to answer''
reportable, ablatable, and regression-testable for the first time. IIRC
questions are written by humans who cannot see the linked documents, so a
substantial fraction is unanswerable in context (the \emph{none} class). From
the official dev split (1{,}301 questions) we stratify by answer\_type (seed
88) into a frozen held-out set ($n=130$: span 59 / value 23 / binary 13 /
none 35) and a tune pool (remaining 1{,}171; no model weights trained). The
held-out set is tested once under the Phase-1 strict exact-only
lattice-direct protocol---38 LLM calls, all query-side---and that run is the
final score.

\begin{table}[t]
  \centering
  \caption{IIRC main results (held-out $n=130$; dev\_tune as drift check).}
  \label{tab:iirc_main}
  \footnotesize
  \begin{tabular}{lll}
    \toprule
    Metric & Held-out 130 & dev\_tune 1{,}171 \\
    \midrule
    EM$_{\mathrm{full}}$ (Phase-1-hit denominator) & 0.0417 (1/24) & 0.0264 (6/227) \\
    Recall measure (answerable EM) & 0.0105 & 0.0070 \\
    Resolved (Phase-1 hit rate) & 25.3\% (24/95) & 26.4\% \\
    Leak rate (none answered) & 0.029 (1/35) & 0.016 (5/312) \\
    Abstain accuracy & 0.971 (34/35) & 0.984 \\
    Precision measure (abstain $\times$ non-leak) & \textbf{0.9437} & 0.9682 \\
    \bottomrule
  \end{tabular}
\end{table}



Held-out and tune pool agree item by item
(Table~\ref{tab:iirc_main}): zero drift. The single leak is a Phase-1
value-match false positive, not generative hallucination. The 35 unanswerable
questions split into 21 no\_hit, 13 sentinel abstains, and 1 leak; the 93
answerable abstains split into 71 no\_hit (39 in-grid / 32 not-in-grid) and 22
sentinel refusals (two-axis attribution: phase $\times$ gold-in-grid,
Appendix Table~\ref{tab:iirc_attribution}).

\subsection{Value Cleaning (Ablation)}
\label{sec:ablation}

A dev-side ablation (IIRC dev $n=1{,}301$, Phase-1 only, same pipeline;
details in Appendix~\ref{app:cleaning}) rewrites sentence-level dimension
values into clean entities on the ingestion side: EM$_{\mathrm{full}}$ rises
0.0156 $\to$ 0.0279 (+79\%) and the leak rate falls 0.052 $\to$ 0.017
($-$67\%)---the leak reduction far exceeds the recall gain, a payoff only the
two-measure separation can characterize.

\subsection{Two Honest Boundaries}
\label{sec:boundaries}

(a)~\emph{value boundary} (IIRC dev 233 questions; held-out 23 all abstained
or wrong): answers are 100\% derived quantities whose anchors exist in the
lattice only $\sim$1.5\% of the time---abstaining \emph{is} correct (echoing
Section~\ref{sec:operators}). (b)~\emph{binary boundary} (IIRC dev 128
questions; held-out 13): the compared entities are mostly not lattice
subjects, so Boolean comparison cannot execute---compare degrades to
abstain, its fourth form (\S\ref{sec:operators}). Low EM and high abstain
accuracy are two faces of one boundary: with no anchor in the lattice, not
answering is what the dual-measure framework rewards.
\section{Conclusion}

We presented \LatWeave{}: multi-hop QA compiled into three deterministic
lattice operators---\emph{meet}, \emph{compare}, \emph{abstain}---with LLMs
confined to construction and query planning; the answer path is zero-LLM,
zero-task-training, and auditable end to end. Across six public benchmarks,
deterministic execution pays no penalty inside its operating envelope, and
failure ends in honest abstention. The method has been tested in an actual
production environment. Deterministic reasoning is not meant to
replace LLMs, but to lift the answer path from a probabilistic black box into
an auditable white box.

\bibliographystyle{ACM-Reference-Format}
\bibliography{references}

@inproceedings{lewis2020retrieval,
  title={Retrieval-augmented generation for knowledge-intensive {NLP} tasks},
  author={Lewis, Patrick and Perez, Ethan and Piktus, Aleksandra and Petroni, Fabio and Karpukhin, Vladimir and Goyal, Naman and K{\"u}ttler, Heinrich and Lewis, Mike and Yih, Wen-tau and Rockt{\"a}schel, Tim and others},
  booktitle={Advances in Neural Information Processing Systems (NeurIPS)},
  volume={33},
  pages={9459--9474},
  year={2020}
}

@article{edge2024from,
  title={From local to global: A graph {RAG} approach to query-focused summarization},
  author={Edge, Darren and Trinh, Ha and Cheng, Nathan and Bradley, Joshua and Chao, Alex and Mody, Avik and Truitt, Steven and Larson, Jonathan},
  journal={arXiv preprint arXiv:2404.16130},
  year={2024}
}

@inproceedings{gutierrez2025hipporag2,
  title={From {RAG} to memory: Non-parametric continual learning for large language models},
  author={Guti{\'e}rrez, Bernal Jim{\'e}nez and Shu, Yiheng and Qi, Weijian and Zhou, Sizhe and Su, Yu},
  booktitle={Proceedings of the 42nd International Conference on Machine Learning (ICML)},
  year={2025},
  note={arXiv:2502.14802}
}

@inproceedings{ko2025cooprag,
  title={Cooperative retrieval-augmented generation for question answering: Mutual information exchange and ranking by contrasting layers},
  author={Ko, Youmin and Seo, Sungjong and Kim, Hyunjoon},
  booktitle={Advances in Neural Information Processing Systems 39 (NeurIPS)},
  year={2025},
  note={arXiv:2512.10422; Appendix C.2 reproduces the full EM/F1 table for the GraphRAG/LightRAG/HippoRAG family under the HippoRAG 2 protocol}
}

@inproceedings{xiong2021mdr,
  title={Answering Complex Open-domain Questions with Multi-hop Dense Retrieval},
  author={Xiong, Wenhan and Li, Xiang Lorraine and Iyer, Srini and Du, Jingfei and Lewis, Patrick and Wang, William Yang and Mehdad, Yashar and Yih, Wen-tau and Riedel, Sebastian and Kiela, Douwe and O{\u{g}}uz, Barlas},
  booktitle={Proceedings of the Ninth International Conference on Learning Representations (ICLR 2021)},
  year={2021}
}

@article{guo2024lightrag,
  title={{LightRAG}: Simple and Fast Retrieval-Augmented Generation},
  author={Guo, Zirui and Xia, Lianghao and Yu, Yanhua and Ao, Tu and Huang, Chao},
  journal={arXiv preprint arXiv:2410.05779},
  year={2024}
}

@inproceedings{sun2024thinkongraph,
  title={Think-on-Graph: Deep and Responsible Reasoning of Large Language Model on Knowledge Graph},
  author={Sun, Jiashuo and Xu, Chengjin and Tang, Lumingyuan and Wang, Saizhuo and Lin, Chen and Gong, Yeyun and Ni, Lionel M. and Shum, Heung-Yeung and Guo, Jian},
  booktitle={Proceedings of the Twelfth International Conference on Learning Representations (ICLR 2024)},
  year={2024}
}

@inproceedings{jiang2023flare,
  title={Active Retrieval Augmented Generation},
  author={Jiang, Zhengbao and Xu, Frank F. and Gao, Luyu and Sun, Zhiqing and Liu, Qian and Dwivedi-Yu, Jane and Yang, Yiming and Callan, Jamie and Neubig, Graham},
  booktitle={Proceedings of the 2023 Conference on Empirical Methods in Natural Language Processing (EMNLP)},
  pages={7969--7992},
  year={2023}
}

@inproceedings{yang2018hotpot,
  title={{HotpotQA}: A dataset for diverse, explainable multi-hop question answering},
  author={Yang, Zhilin and Peng, Pengqi and Dai, Zhen and Hu, Yanxia and Carbonell, Jaime and Salakhutdinov, Ruslan},
  booktitle={Proceedings of the 2018 Conference on Empirical Methods in Natural Language Processing (EMNLP)},
  pages={2369--2380},
  year={2018},
  doi={10.18653/v1/D18-1259}
}

@inproceedings{ho2020constructing,
  title={Constructing A Multi-hop {QA} Dataset for Comprehensive Evaluation of Reasoning Steps},
  author={Ho, Xanh and Duong Nguyen, Anh-Khoa and Sugawara, Saku and Aizawa, Akiko},
  booktitle={Proceedings of the 28th International Conference on Computational Linguistics (COLING)},
  pages={6609--6625},
  year={2020}
}

@inproceedings{wille1982restructuring,
  title={Restructuring lattice theory: An approach based on hierarchies of concepts},
  author={Wille, Rudolf},
  booktitle={Ordered Sets},
  pages={445--470},
  year={1982},
  publisher={Springer},
  doi={10.1007/978-94-009-7798-3\_15}
}

@book{davey2002introduction,
  title={An Introduction to Lattices and Order},
  author={Davey, Brian A. and Priestley, Hilary A.},
  year={2002},
  edition={2nd},
  publisher={Cambridge University Press},
  doi={10.1017/CBO9780511809088}
}

@article{dilworth1950,
  title={A decomposition theorem for partially ordered sets},
  author={Dilworth, Robert P.},
  journal={Annals of Mathematics},
  volume={51},
  number={1},
  pages={161--166},
  year={1950}
}

@inproceedings{zhang2024beamretrieval,
  title={End-to-End Beam Retrieval for Multi-Hop Question Answering},
  author={Zhang, Jiahao and Zhang, Haiyang and Zhang, Dongmei and Liu, Yong and Huang, Shen},
  booktitle={Proceedings of the 2024 Conference of the North American Chapter of the Association for Computational Linguistics (NAACL)},
  pages={1718--1731},
  year={2024}
}

@inproceedings{islam2024openrag,
  title={{OPEN-RAG}: Enhanced Retrieval-Augmented Reasoning with Open-Source Large Language Models},
  author={Islam, Shayekh Bin and Rahman, Md Asib and Hossain, K S M Tozammel and Hoque, Enamul and Joty, Shafiq and Parvez, Md Rizwan},
  booktitle={Findings of the Association for Computational Linguistics: EMNLP 2024},
  pages={14231--14244},
  year={2024},
  publisher={Association for Computational Linguistics},
  address={Miami, Florida, USA},
  doi={10.18653/v1/2024.findings-emnlp.831},
  note={arXiv:2410.01782}
}

@inproceedings{asai2024selfrag,
  title={{Self-RAG}: Learning to Retrieve, Generate, and Critique through Self-Reflection},
  author={Asai, Akari and Wu, Zeqiu and Wang, Yizhong and Sil, Avirup and Hajishirzi, Hannaneh},
  booktitle={Proceedings of the Twelfth International Conference on Learning Representations (ICLR)},
  year={2024}
}

@inproceedings{trivedi2023ircot,
  title={Interleaving Retrieval with Chain-of-Thought Reasoning for Knowledge-Intensive Multi-Step Questions},
  author={Trivedi, Harsh and Balasubramanian, Niranjan and Khot, Tushar and Sabharwal, Ashish},
  booktitle={Proceedings of the 61st Annual Meeting of the Association for Computational Linguistics (ACL)},
  pages={10014--10037},
  year={2023},
  doi={10.18653/v1/2023.acl-long.557}
}

@inproceedings{liu2025hoprag,
  title={{HopRAG}: Multi-Hop Reasoning for Logic-Aware Retrieval-Augmented Generation},
  author={Liu, Hao and Wang, Zhengren and Chen, Xi and Li, Zhiyu and Xiong, Feiyu and Yu, Qinhan and Zhang, Wentao},
  booktitle={Findings of the Association for Computational Linguistics: ACL 2025},
  pages={1897--1913},
  year={2025},
  address={Vienna, Austria},
  publisher={Association for Computational Linguistics},
  doi={10.18653/v1/2025.findings-acl.97}
}

@inproceedings{trivedi2022musique,
  title={{MuSiQue}: Multihop Question Answering via Single-Hop Question Composition},
  author={Trivedi, Harsh and Balasubramanian, Niranjan and Khot, Tushar and Sabharwal, Ashish},
  booktitle={Transactions of the Association for Computational Linguistics (TACL)},
  volume={10},
  pages={539--554},
  year={2022},
  publisher={MIT Press}
}

@inproceedings{krishna2025frames,
  title={Fact, Fetch, and Reason: A Unified Evaluation of Retrieval-Augmented Generation},
  author={Krishna, Satyapriya and Krishna, Kalpesh and Mohananey, Anhad and Schwarcz, Steven and Stambler, Adam and Upadhyay, Shyam and Faruqui, Manaal},
  booktitle={Proceedings of the 2025 Conference of the North American Chapter of the Association for Computational Linguistics: Human Language Technologies (Volume 1: Long Papers)},
  pages={4745--4759},
  year={2025},
  address={Albuquerque, New Mexico},
  publisher={Association for Computational Linguistics},
  doi={10.18653/v1/2025.naacl-long.243}
}

@inproceedings{ferguson2020iirc,
  title={{IIRC}: A Dataset of Incomplete Information Reading Comprehension Questions},
  author={Ferguson, James and Gardner, Matt and Hajishirzi, Hannaneh and Khot, Tushar and Dasigi, Pradeep},
  booktitle={Proceedings of the 2020 Conference on Empirical Methods in Natural Language Processing (EMNLP)},
  pages={1137--1147},
  year={2020},
  address={Online},
  publisher={Association for Computational Linguistics},
  doi={10.18653/v1/2020.emnlp-main.86}
}

@article{ji2023survey,
  title={Survey of hallucination in natural language generation},
  author={Ji, Ziwei and Lee, Nayeon and Frieske, Rita and Yu, Tiezheng and Su, Dan and Xu, Yan and Ishii, Etsuko and Bang, Ye Jin and Madotto, Andrea and Fung, Pascale},
  journal={ACM Computing Surveys},
  volume={55},
  number={12},
  pages={1--38},
  year={2023},
  doi={10.1145/3571730}
}

@inproceedings{kalai2024calibrated,
  title={Calibrated Language Models Must Hallucinate},
  author={Kalai, Adam Tauman and Vempala, Santosh S.},
  booktitle={Proceedings of the 56th Annual ACM Symposium on Theory of Computing (STOC 2024)},
  pages={160--171},
  year={2024},
  publisher={ACM},
  doi={10.1145/3618260.3649777}
}

@book{ganter2012formal,
  title={Formal Concept Analysis: Mathematical Foundations},
  author={Ganter, Bernhard and Wille, Rudolf},
  year={1999},
  publisher={Springer},
  doi={10.1007/978-3-642-59830-2}
}

@inproceedings{zhang2018variational,
  title={Variational reasoning for question answering with knowledge graph},
  author={Zhang, Yuyu and Dai, Hanjun and Kozareva, Zornitsa and Smola, Alexander and Song, Le},
  booktitle={Proceedings of the Thirty-Second AAAI Conference on Artificial Intelligence (AAAI)},
  pages={6069--6076},
  year={2018}
}

@inproceedings{miller2016key,
  title={Key-value memory networks for directly reading documents},
  author={Miller, Alexander and Fisch, Adam and Dodge, Jesse and Karimi, Amir-Hossein and Bordes, Antoine and Weston, Jason},
  booktitle={Proceedings of the 2016 Conference on Empirical Methods in Natural Language Processing (EMNLP)},
  pages={1400--1409},
  year={2016}
}

@inproceedings{sun2018open,
  title={Open domain question answering using early fusion of knowledge bases and text},
  author={Sun, Haitian and Dhingra, Bhuwan and Zaheer, Manzil and Mazaitis, Kathryn and Salakhutdinov, Ruslan and Cohen, William},
  booktitle={Proceedings of the 2018 Conference on Empirical Methods in Natural Language Processing (EMNLP)},
  pages={4231--4242},
  year={2018}
}

@inproceedings{sun2019pullnet,
  title={Pull{Net}: Open domain question answering with iterative retrieval on knowledge bases and text},
  author={Sun, Haitian and Bedrax-Weiss, Tania and Cohen, William},
  booktitle={Proceedings of the 2019 Conference on Empirical Methods in Natural Language Processing (EMNLP)},
  pages={2380--2390},
  year={2019}
}

@inproceedings{saxena2020improving,
  title={Improving multi-hop question answering over knowledge graphs using knowledge base embeddings},
  author={Saxena, Apoorv and Tripathi, Aditay and Talukdar, Partha},
  booktitle={Proceedings of the 58th Annual Meeting of the Association for Computational Linguistics (ACL)},
  pages={4498--4507},
  year={2020}
}

@inproceedings{he2021improving,
  title={Improving multi-hop knowledge base question answering by learning intermediate supervision signals},
  author={He, Gaole and Lan, Yunshi and Jiang, Jing and Zhao, Wayne Xin and Wen, Ji-Rong},
  booktitle={Proceedings of the 14th ACM International Conference on Web Search and Data Mining (WSDM)},
  pages={553--561},
  year={2021}
}

@inproceedings{qiu2020stepwise,
  title={Stepwise reasoning for multi-relation question answering over knowledge graph with weak supervision},
  author={Qiu, Yunqi and Wang, Yuanzhuo and Jin, Xiaolong and Zhang, Kun},
  booktitle={Proceedings of the 13th International Conference on Web Search and Data Mining (WSDM)},
  pages={474--482},
  year={2020}
}

@inproceedings{luo2024reasoning,
  title={Reasoning on graphs: Faithful and interpretable large language model reasoning},
  author={Luo, Linhao and Li, Yuan-Fang and Haffari, Gholamreza and Pan, Shirui},
  booktitle={Proceedings of the Twelfth International Conference on Learning Representations (ICLR)},
  year={2024}
}

@inproceedings{fu2026spathrag,
  title={{S-Path-RAG}: Semantic-aware shortest-path retrieval augmented generation for multi-hop knowledge graph question answering},
  author={Fu, Rong and Wang, Yemin and Xu, Tianxiang and Liu, Yongtai and Tang, Weizhi and Wu, Wangyu and Ma, Xiaowen and Fong, Simon},
  booktitle={Proceedings of the ACM Web Conference 2026 (WWW)},
  pages={4057--4068},
  year={2026},
  doi={10.1145/3774904.3792459},
  note={arXiv:2603.23512}
}

@inproceedings{yang2026compactrag,
  title={{CompactRAG}: Reducing {LLM} calls and token overhead in multi-hop question answering},
  author={Yang, Hao and Wei, Wei and Yang, Zhiyu and Zhang, Xupeng and Zhang, Yunjie and Yang, Lin},
  booktitle={Proceedings of the ACM Web Conference 2026 (WWW)},
  pages={2240--2251},
  year={2026},
  doi={10.1145/3774904.3792512},
  note={arXiv:2602.05728}
}

@inproceedings{han2026reasoning,
  title={Reasoning by Exploration: A Unified Approach to Retrieval and Generation over Graphs},
  author={Han, Haoyu and Guo, Kai and Shomer, Harry and Wang, Yu and Chu, Yucheng and Li, Hang and Ma, Li and Tang, Jiliang},
  booktitle={Proceedings of the ACM Web Conference 2026 (WWW)},
  year={2026},
  note={arXiv:2510.07484}
}

@inproceedings{lien2026hyperrag,
  title={{HyperRAG}: Reasoning {N}-ary Facts over Hypergraphs for Retrieval Augmented Generation},
  author={Lien, Wen-Sheng and Chan, Yu-Kai and Hsiao, Hao-Lung and Ruan, Bo-Kai and Chiang, Meng-Fen and Chen, Chien-An and Yeh, Yi-Ren and Shuai, Hong-Han},
  booktitle={Proceedings of the ACM Web Conference 2026 (WWW)},
  pages={2465--2476},
  year={2026},
  doi={10.1145/3774904.3792710},
  note={arXiv:2602.14470}
}

@article{elyaniv2010foundations,
  title={On the foundations of noise-free selective classification},
  author={El-Yaniv, Ran and Wiener, Yair},
  journal={Journal of Machine Learning Research (JMLR)},
  volume={11},
  number={53},
  pages={1605--1641},
  year={2010}
}

@inproceedings{geifman2017selective,
  title={Selective classification for deep neural networks},
  author={Geifman, Yonatan and El-Yaniv, Ran},
  booktitle={Advances in Neural Information Processing Systems (NeurIPS)},
  volume={30},
  year={2017},
  note={arXiv:1705.08500}
}

@inproceedings{kamath2020selective,
  title={Selective question answering under domain shift},
  author={Kamath, Amita and Jia, Robin and Liang, Percy},
  booktitle={Proceedings of the 58th Annual Meeting of the Association for Computational Linguistics (ACL)},
  pages={5684--5696},
  year={2020},
  doi={10.18653/v1/2020.acl-main.503}
}

@inproceedings{lin2022truthfulqa,
  title={{TruthfulQA}: Measuring how models mimic human falsehoods},
  author={Lin, Stephanie and Hilton, Jacob and Evans, Owain},
  booktitle={Proceedings of the 60th Annual Meeting of the Association for Computational Linguistics (Volume 1: Long Papers)},
  pages={3214--3252},
  year={2022},
  doi={10.18653/v1/2022.acl-long.229}
}

@misc{llmplanning2025kgqa,
  title={Efficient Multi-Hop Question Answering over Knowledge Graphs via {LLM} Planning and Embedding-Guided Search},
  author={Shrestha, Manil and Kim, Edward},
  year={2025},
  eprint={2511.19648},
  archivePrefix={arXiv},
  primaryClass={cs.CL}
}

@inproceedings{whitehead2022reliable,
  title={Reliable Visual Question Answering: Abstain Rather Than Answer Incorrectly},
  author={Whitehead, Spencer and Petryk, Suzanne and Shakib, Vedaad and Gonzalez, Joseph and Darrell, Trevor and Rohrbach, Anna and Rohrbach, Marcus},
  booktitle={Computer Vision -- ECCV 2022, Part XXXVI},
  series={Lecture Notes in Computer Science},
  volume={13696},
  pages={148--166},
  publisher={Springer},
  year={2022},
  doi={10.1007/978-3-031-20059-5\_9},
  note={arXiv:2204.13631}
}

@inproceedings{bai2026autoschemakg,
  title={AutoSchema{KG}: Autonomous knowledge graph construction through dynamic schema induction from web-scale corpora},
  author={Bai, Jiaxin and Fan, Wei and Hu, Qi and Zong, Qing and Li, Chunyang and Tsang, Hong Ting and others},
  booktitle={Proceedings of the 64th Annual Meeting of the Association for Computational Linguistics (ACL)},
  year={2026},
  note={arXiv:2505.23628}
}

@inproceedings{angelopoulos2023conformal,
  title={Conformal risk control},
  author={Angelopoulos, Anastasios N. and Bates, Stephen and Cand{\`e}s, Emmanuel J. and Jordan, Michael I. and Lei, Lihua},
  booktitle={Proceedings of Machine Learning Research (PMLR)},
  volume={202},
  pages={2610--2639},
  year={2023},
  note={arXiv:2208.02814}
}

@inproceedings{liang2025kag,
  author={Liang, Lei and Bo, Zhongpu and Gui, Zhengke and Zhu, Zhongshu and Zhong, Ling and Zhao, Peilong and Sun, Mengshu and Zhang, Zhiqiang and Zhou, Jun and Chen, Wenguang and Zhang, Wen and Chen, Huajun},
  title={KAG: Boosting {LLMs} in Professional Domains via Knowledge Augmented Generation},
  booktitle={Companion Proceedings of the ACM on Web Conference 2025},
  pages={334--343},
  year={2025},
  doi={10.1145/3701716.3715240}
}

@inproceedings{ren2026whentotrust,
  title={When to Trust: A Causality-Aware Calibration Framework for Accurate {KG}-{RAG}},
  author={Ren, Jing and Li, Bowen and Xu, Ziqi and Zhang, Xikun and Fayek, Haytham and Li, Xiaodong},
  booktitle={Proceedings of The Web Conference (WWW)},
  year={2026},
  note={arXiv:2601.09241}
}

\appendix
\section{FRAMES: Controlled Single-Variable Ablation---Locating the Failing Stage}
\label{app:frames}

FRAMES (Krishna et al., 2025; factuality $\times$ multi-hop, 824 questions,
each with gold evidence pages) is used here not as an additional, harder
stress test but as a controlled single-variable ablation that separates two
stages---whether the seed anchor lands in the lattice, and whether the chain
semantics line up---to locate which stage the chain break actually occurs
in. We apply a uniform 90/10 development/frozen split (seed 88): dev 742
questions for development, held-out 82 frozen and used for final testing only;
construction is question-blind open extraction, with the LLM used only for
query-side planning.

Main results: dev EM $=$ 19/742 (2.56\%), oracle diagnostic ceiling (gold
falls in the candidate set; not the main EM measure) 36/742 (4.9\%). The
held-out 82 are reported under two settings: evidence-page-excluded
construction gives EM $=$ 1/82 (F1 $=$ 0.0521), evidence-page-included
construction gives EM $=$ 1/82 (F1 $=$ 0.1826)---the two EMs are identical
(statistically inseparable), yet including the evidence pages lifts
seed-entity resolution from 20.7\% to 100\% and the share of chain-walkable
questions from 15\% to 87\%. This is the controlled single-variable
conclusion of this appendix: adding the pages rescues the first stage (``does
the anchor land''), while the bottleneck at the second stage (``does the chain
line up'') is unchanged---so the failing stage is precisely located in chain
semantics, not in extraction coverage.

Three-line falsification chain: (1)~after executor-layer fixes EM is flat, a
net effect of zero; (2)~a replanning pilot scores 0 hits on 15 questions ---
when the break is a coverage gap there is no alternative chain to take;
(3)~dissecting 13 questions from the targeted re-extraction candidate list, 12
have an intermediate entity that is simply not in the lattice (never landed as
a lattice point), while in 4 control cases (where the entity has 64--129
dimensions in the lattice) the break-point relation still does not exist in
its instance set---the break comes from the planner inventing a relation the
entity does not have, a chain-design error against which re-extraction is
ineffective in principle. The three lines jointly support
Section~\ref{sec:grounding}.

Capability strata: FACT\_MEET / BOOL / EXTREME / COMPUTE, with the decision
table in Appendix~\ref{app:protocol}. Transferable conclusion: filling
coverage does not equal getting chains right---construction coverage is only
a necessary condition, whereas chain-semantic alignment and chain
landing-point precision (whether gold enters the returned set) are the
bottleneck on the sufficient-condition side (in the same direction as the
HotpotQA H2 oracle injection experiment).

\section{Evaluation Protocol Details}
\label{app:protocol}

Measures: EM/F1 follow each benchmark's official evaluation script,
and F1* is this paper's variant (defined in Section~\ref{sec:metrics}). Seed
88 throughout; the 90/10 split manifests are frozen as files (each benchmark's
held-out manifest is on disk). LLM configuration: temp 0.0, context\_size 32K
(configurable), max\_tokens 1280 (with automatic escalation to 2048 on
truncation). The FRAMES capability-class decision table (FACT\_MEET / BOOL /
EXTREME / COMPUTE) is included in the supplementary material.

\section{Prompt Templates}
\label{app:prompts}

The extraction prompt (hybrid extraction, A/B modes), the planner prompt
(including the KNOWN RELATIONS injection and the entity-menu variant), and the
RAG baseline reader prompt are provided in full in the supplementary
material.

\section{Construction Statistics}
\label{app:stats}

Per-domain scale (frozen final values): the 2Wiki dataset has 1.2M instances
(125{,}760 passages); MetaQA has 19 dimensions / 269{,}482 instances
(134{,}741 triples / 43{,}234 entities); the full MuSiQue lattice has 40{,}566
instances / 1{,}741 dimensions, and the held-out-only sublattice has 3{,}996
instances / 535 passages / 502 dimensions / 5{,}202 nodes; HotpotQA has 852
dimensions / 120{,}981 instances / 26{,}258 subject nodes; IIRC has 350
normalized dimensions / 6{,}768 instances.

\section{HotpotQA Re-extraction Repair, Measured}
\label{app:reextract}

Can re-extracting the same corpus at the build layer (same open extractor)
repair the coverage gap? Two small-scale experiments answer no. Hop-1
($n=150$): only 16 questions (10.7\%) gain a genuine first-hop bridge in the
lattice (an edge connecting the seed to the gold intermediate entity); and
connectivity is not correctness---replaying all formerly hop-1-broken
questions ($n=1{,}509$) under the frozen configuration (re-extraction,
relation mapping, and planner fixes included), the first-hop connectivity
rate is 78.9\%, 21.9\% of connected questions are ultimately answered
correctly (51.3\% resolved), and the bucket's EM rises from 0 to 17.3\% ---
the repair works but has a clear ceiling; the rest either break at later hops
or the answer never enters the returned set (landing points and value
granularity). Extrapolating with the measured 10.7\% bridge-recovery rate and
21.9\% conversion, even re-extracting all 1,613 first-hop gaps would add only
about +0.6pp to full-pool (dev 6,664) EM, optimistically. Hop-2 ($n=36$):
conversion from re-extraction to a fully correct answer is only 5.6\% (2/36;
small $n$, 95\% confidence interval roughly 0.7\%--18.9\%). A per-question
residual decomposition (diagnostic statistics on the dev tune pool
$n=6{,}664$, held-out untouched) rewrites the bottleneck from ``choosing
wrong among candidates'' to ``gold not in the returned set'': among
resolved-but-wrong questions ($n=1{,}437$), 77\% have their gold answer
outside the system's returned candidate set at all---landing-point drift,
coverage gaps, or extraction granularity; only about 11\% have the gold's
content words fully inside the returned set (mostly as noisy long strings).
This 77\% is the residual after three repair paths---executor fixes
(net-zero effect), per-hop replanning (0 of 15 hits), and re-extraction ---
have all been rejected; its magnitude corresponds to the true scale of the
coverage gap on boundary datasets (Appendix Table~\ref{tab:attribution}), not
an artifact of the evaluation protocol. Questions whose returned set contains
two or more candidates account for only 9.3\% (averaging just 3.6
candidates), and since the primary metric is any-hit (gold in the returned
set counts as correct, Section~\ref{sec:metrics}), ``choosing the wrong value
among candidates'' affects only a small share.

\section{Value Cleaning Ablation (IIRC dev)}
\label{app:cleaning}

An early version of open extraction poured whole sentences into dimension
values (value $=$ sentence), causing two kinds of contamination: on answerable
questions value matching failed and depressed the recall measure, and on
unanswerable questions sentence-level values were partially matched and
triggered leaks. Value cleaning rewrites sentence values of entity-typed
dimensions into clean entities on the ingestion side only, with no change to
query or lattice logic: EM$_{\mathrm{full}}$ rises from 0.0156 to 0.0279
(+79\%), the leak rate falls from 0.052 to 0.017 ($-$67\%), abstain accuracy
rises from 0.948 to 0.983, and the precision measure rises from 0.899 to 0.966.
The leak reduction far exceeds the recall gain---a payoff pattern that only
the separation of the two measures can characterize, and direct evidence of the
framework's diagnostic value.

\section{Additional Results Tables}
\label{app:extra}

Tables~\ref{tab:positioning}--\ref{tab:iirc_attribution} collect the
supporting detail tables referenced from the main text: the approach-family
positioning table, the 2WikiMultihopQA ablation, the MetaQA answer-set-size
bucketing, the cross-dataset failure attribution matrix, and the
per-question IIRC abstention attribution.

\begin{table}[!htbp]
  \centering
  \caption{Positioning: only \LatWeave{} ticks all four columns. Zero-training is graded
  because several families mix trained and training-free members (see footnotes).
  Detection (auditability) is graded because chunk-ID provenance yields engineering-level
  auditability only: the step from retrieved passage to answer remains unverifiable.}
  \label{tab:positioning}
  \footnotesize
  \begin{tabular}{@{}p{2.55cm}cccc@{}}
    \toprule
    Approach family & Zero-train & Determ. & Audit. & Abstain \\
    \midrule
    Neural multi-hop retrievers & partial\textsuperscript{1} & $\times$ & $\times$ & $\times$ \\
    RAG / GraphRAG              & partial\textsuperscript{2} & $\times$ & eng.-level & $\times$ \\
    KBQA / semantic parsing     & $\times$\textsuperscript{3} & \checkmark & \checkmark & $\times$ \\
    Selective prediction        & $\times$\textsuperscript{4} &---& $\times$ & prob.~thresh. \\
    \midrule
    \LatWeave{} (this work)     & \checkmark & \checkmark & \checkmark & \checkmark \\    \bottomrule
  \end{tabular}
  \vspace{1pt}
  \par\footnotesize\raggedright
  \textsuperscript{1}~MDR and Beam Retrieval train in-task retrievers; IRCoT and HopRAG are
  training-free (LLMs in the loop).
  \textsuperscript{2}~Self-RAG and RoG fine-tune; FLARE, GraphRAG, HippoRAG~2 and
  KAG~\cite{liang2025kag} are training-free. KAG further compiles questions into
  logical forms for hybrid reasoning, yet the final answer is still LLM-synthesized.
  Causality-aware recalibration of KG-RAG confidence~\cite{ren2026whentotrust} is
  likewise training-free, yet it calibrates a probabilistic answer generator rather
  than removing one.
  Self-RAG and FLARE gate whether to retrieve, not whether to answer: the model still
  generates an answer, so they provide no abstention mechanism (cf.~Section~2.4, where
  selective prediction does, but via confidence thresholds).
  \textsuperscript{3}~Exception: LLM-planned BFS over a given KG~\cite{llmplanning2025kgqa}
  needs no task training (Section~2.3). Schema-free KG construction from
  text~\cite{bai2026autoschemakg} relaxes the predefined schema, but its query side is
  probabilistic retrieval over the induced graph, not deterministic execution.
  \textsuperscript{4}~Requires a trained selector or held-out calibration; conformal risk
  control calibrates the abstention threshold distribution-free (removing calibration
  drift)~\cite{angelopoulos2023conformal}, yet abstention remains a thresholding
  operation and the underlying answer generator remains probabilistic. Trained
  refusal rules can also be tuned to abstain, but such a refusal is a learned
  probabilistic decision, not a structural guarantee that can be audited
  independently of the model.
\end{table}

\begin{figure}[!htbp]
  \centering
  \includegraphics[width=\linewidth]{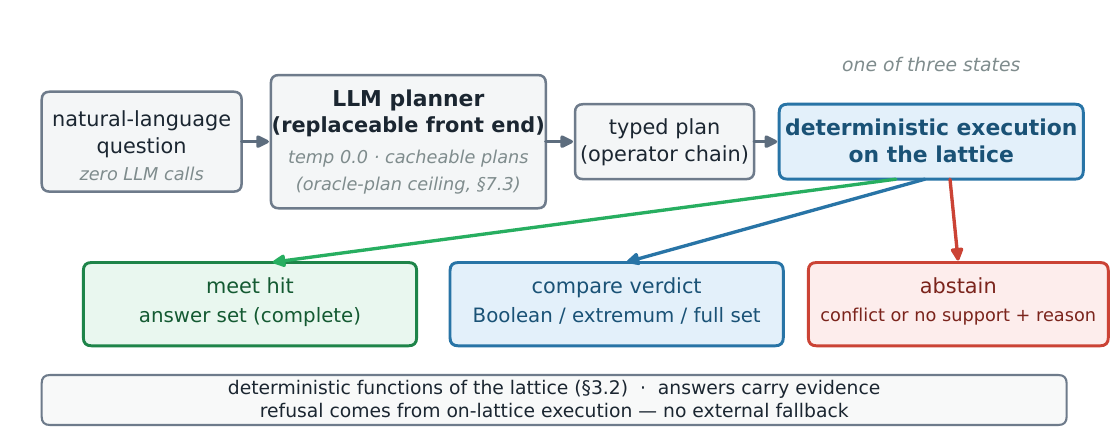}
  \caption{Query-time flow. The LLM planner compiles the question into a
  typed plan; deterministic execution on the lattice returns a \emph{meet}
  hit, a \emph{compare} verdict, or \emph{abstain}; the answer ships with its
  evidence text.}
  \label{fig:pipeline}
\end{figure}


\begin{figure}[!htbp]
  \centering
  \includegraphics[width=\linewidth]{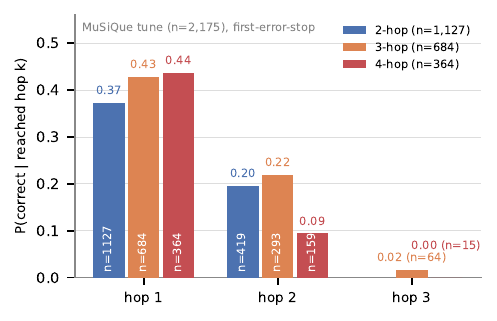}
  \caption{Per-hop conditional meet accuracy $P(\text{correct} \mid \text{chain reaches
  hop } k)$ on the MuSiQue tune pool ($n=2{,}175$), grouped by the question's true
  depth.}
  \label{fig:hopwise}
\end{figure}

\begin{table}[!htbp]
  \centering
  \caption{2WikiMultihopQA ablation (dev tune pool, historical protocol $n=11{,}318$).}
  \label{tab:twowiki_ablation}
  \small
  \begin{tabular}{lcc}
    \toprule
    Configuration & EM & Resolved \\
    \midrule
    Full system & 0.873 & 92.0\% \\
    $-$ train gold transfer (dev-only) & 0.731 & 85.3\% \\
    $-$ alias normalization & 0.853 & 90.8\% \\
    $-$ rule-based re-extraction (E15) & 0.843 & 90.4\% \\
    \bottomrule
  \end{tabular}
  \vspace{1pt}
  \par\footnotesize\raggedright
  The full system, replayed after domain rebuild (grid-point hash dedup,
  1.2M instances), gives tune-pool EM 0.886 / held-out 0.865; the remaining
  rows are historical-protocol ablation values.
\end{table}

\begin{table}[!htbp]
  \centering
  \caption{MetaQA test bucketed by gold answer-set size (completeness structural analysis).}
  \label{tab:metaqa_buckets}
  \small
  \begin{tabular}{lllll}
    \toprule
    $|A^*|$ & $n$ & any-hit & recall (complete) & EM (strict) \\
    \midrule
    1      & 14{,}834 & 0.9974 & 0.9974 & 0.8332 \\
    2      & 5{,}729  & 0.9977 & 0.9977 & 0.7413 \\
    3--5   & 7{,}277  & 0.9974 & 0.9968 & 0.6442 \\
    6--10  & 4{,}041  & 0.9968 & 0.9968 & 0.5494 \\
    $\geq$11 & 7{,}212 & 0.9978 & 0.9976 & 0.3788 \\
    \midrule
    All    & 39{,}093 & 0.9975 & 0.9973 & 0.6714 \\
    \bottomrule
  \end{tabular}
\end{table}

\begin{table}[!htbp]
  \centering
  \caption{Failure attribution matrix.}
  \label{tab:attribution}
  \footnotesize
  \setlength{\tabcolsep}{3pt}
  \begin{tabular}{@{}p{1.8cm}p{3.1cm}p{2.2cm}@{}}
    \toprule
    Dataset & Attribution & Count (share) \\
    \midrule
    HotpotQA & Extraction coverage gap & 667/736 (90.6\%) \\
    (residual broken chains) & \quad H1 missing / H2 missing / seed no-out-edge & 527 / 81 / 59 \\
             & Query-layer planner & 69/736 (9.4\%) \\
    \midrule
    MuSiQue & Break at hop 1 & 117/242 (48.3\%) \\
    (chain-break locus) & Break at hop 2 & 81/242 (33.5\%) \\
             & Break at hop 3 / hop 4 & 16 / 4 (6.6\% / 1.7\%) \\
             & Chain survives (resolved-no-break: 6 correct / 18 wrong) & 24/242 (9.9\%) \\
    \midrule
    FRAMES (dev $n=742$) & Coverage gap compounded by chain-design error & 12/13 dissected: intermediate entity off-lattice \\
    \bottomrule
  \end{tabular}
\end{table}

{\par\vspace{6pt}
\captionof{table}{IIRC held-out abstention attribution on two axes (phase $\times$
gold-in-grid), per-question reproducible ($n=130$). The in-grid probe applies
only to answerable questions---unanswerable ones have no gold answer by
construction. The 22 answerable sentinel abstains split evenly (11 in-grid /
11 not-in-grid). The broken-chain bucket is empty by design on IIRC: this arm
is Phase-1 exact-only single-value mapping with no multi-hop chain to break
(chain-break attribution belongs to Section~\ref{sec:attribution}). A
dev-scale isomorphic analysis (954 answerable questions, older grid convention)
gives the same proportional structure.}
\label{tab:iirc_attribution}
\centering\footnotesize
\setlength{\tabcolsep}{3pt}
\begin{tabular}{@{}llrr@{}}
    \toprule
    & & \multicolumn{2}{c}{Question class} \\
    \cmidrule(lr){3-4}
    Phase & Disposition & None ($n{=}35$) & Answerable ($n{=}95$) \\
    \midrule
    \multirow{3}{*}{Phase-1 hit}
      & answered correct &---& 1 \\
      & answered wrong &---& 1 \\
      & sentinel abstain & 13 & 22 \\
    \midrule
    \multirow{3}{*}{no\_hit}
      & gold in grid (``not met'') & n/a & 39 \\
      & gold not in grid (``never extracted'') & n/a & 32 \\
      & subtotal & 21 & 71 \\
    \midrule
    \multicolumn{2}{l}{Leak (vacuous pass)} & 1 &---\\
    \midrule
    \multicolumn{2}{l}{Total} & 35 & 95 \\
    \bottomrule
\end{tabular}
\par\vspace{6pt}}

\section{Auditability and Material Availability}
\label{app:repro}

To ensure reproducibility and auditability, we provide, with the supplementary
material, all prompt templates (Appendix~\ref{app:prompts}), the held-out split
manifests, per-question evaluation details, and the failure-attribution probe
results, so that the numbers on all six benchmarks can be independently
recomputed under this paper's evaluation protocol; because the system implementation
involves industrial deployment, the source code is not part of the
supplementary material.

\end{document}